\documentclass{article}
\usepackage[utf8]{inputenc}
\usepackage[english]{babel}
\usepackage{graphicx}
\usepackage[ruled,vlined]{algorithm2e}
\usepackage{subcaption}
\usepackage[numbers]{natbib}
\usepackage{multicol}
\usepackage{comment}
\usepackage[left=40mm, right=20mm, top=20mm, bottom=20mm]{geometry}

\title{POSLAN: Disentangling Chat with Positional and Language encoded Post Embeddings}
\author{Bhashithe Abeysinghe\thanks{tabeysinghe2@student.gsu.edu} \and Dhara Shah \and Chris Freas \and Robert Harrison \and Rajshekhar Sunderraman\\\\\small{Department of Computer Science,}\\\small{Georgia State University,}\\\small{Atlanta}}
\date{}
\begin{document}
\maketitle

\begin{abstract}
    Most online message threads inherently will be cluttered and any new user or an existing user visiting after a hiatus will have a difficult time understanding whats being discussed in the thread. Similarly cluttered responses in a message thread makes analyzing the messages a difficult problem. The need for disentangling the clutter is much higher when the platform where the discussion is taking place does not provide functions to retrieve reply relations of the messages. This introduces an interesting problem to which \cite{wang2011learning} phrases as a structural learning problem. We create vector embeddings for posts in a thread so that it captures both linguistic and positional features in relation to a context of where a given message is in. Using these embeddings for posts we compute a similarity based connectivity matrix which then converted into a graph. After employing a pruning mechanisms the resultant graph can be used to discover the reply relation for the posts in the thread. The process of discovering or disentangling chat is kept as an unsupervised mechanism. We present our experimental results on a data set obtained from Telegram\footnote{https://telegram.org/} with limited meta data.
\end{abstract}

\section{Introduction}
Online discussion boards provide a versatile platform which allows users to have fluid and multithreaded discussions about any topic. Examples of discussion boards include Twitter, What's app, and Telegram. Telegram, in particular, is widely used by members of criminal and quasi-criminal Dark Net groups to exchange information and negotiate exchange of goods and funds. Thus, automated scanning, processing, and comprehension of Telegram conversations is a significant problem in understanding and preventing cybercrime.

When a new user joins the conversation, there may be thousands of existing messages and different topics already in the discussion. While human users can readily assimilate the chat history and integrate into the discussion, this is a difficult problem in natural language processing and AI. The conversations are asynchronous and at least somewhat overlapping. Therefore understanding the conversation requires disentangling the different threads so that individual conversations can be analysed. ``Reply-to'' and similar metainformation is useful, but not sufficient, for automatic disentanglement.

Reply to feature can come in different forms; one where the user can select the post to which they are replying to, this will show the whole or a part of the original message in the replying users message. Some discussion boards maintain the reply to feature as a tree like hierarchy where one could look into the thread in a much more informed manner. In any of these discussion boards, it is to be understood that users may or may not use the said functionality or if one chooses to use the features could use it in an incorrect manner which makes the reply structure also wrong. Not only that there maybe situations where a user wants to reply to many messages in the thread at once, most online platforms does not provide such a feature. Considering these reconstructing the reply structure is an interesting and relevant problem that needs to be explored. This problem was phrased as a structural learning \cite{wang2011learning} problem in natural language processing domain.

\begin{figure}
    \centering
    \includegraphics[height=6cm]{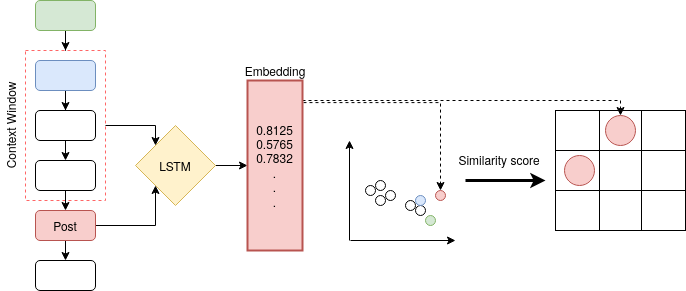}
    \caption{Processing of posts with the context window paradigm 1 and generating the similarity matrix}
    \label{fig:model}
\end{figure}

\section{Problem Statement}
\label{problem_statement}
In this article a `thread' is a linear, chronologically ordered conversation where two or more users take part in a discussion. Problem in simpler terms can be phrased like this; Given a thread $P_i$ compute the reply relations $R$. ($P_i = \{p_{i,1}, p_{i,2}, ..., p_{i,k}\}$), We need to extract post and replies relation ($p_{i_j} \rightarrow p_{i,k}$) for all the posts in the thread. A post ($p_{i,l}$) could have multiple replies, $(p_{i,l} \rightarrow \{p_{i,m}, p_{i,n} ...\}$)

Inspired by the Skipgram model in \cite{mikolov2013distributed} we model our inputs and labels as a singular post ($p_i$) in a thread and the context window of $k$ posts $p_{i-k}, p_{i-k-1}, ..., p_i$ respectively. The context will allow the network to encode positional information with respect to its linguistic features into the embeddings so that not only linguistic features are captured. In a typical conversation context is usually previous questions, answers and comments on a similar theme. If the reply relation is not extracted during the time the conversation is happening (not online) it is also possible to include posts which come after $p_i$ into the context such as, ($p_{i-k}, p_{i-k-1}, ..., p_i, p_{i+1}, p_{i+2}, ..., p_{i+k}$). This gives the network a unique perspective of the conversation which in an online setting would be difficult to incorporate.

Schisms in a conversation can happen due to various reasons \cite{elsner2008you}, while we do not observe all reasons we base that a schism can happen either due to a conversation dying out because users didn't engage in it, or there was a more interesting point or a question raised by a participant. We believe if we can detect schisms, we can efficiently model the post to parent relationship by correctly clustering messages.

To handle first type of schisms we can use the language similarity measure, since the vector embeddings itself incorporates the language with the positional information we can use the same here. The other can be modeled as a Hawkes process. A Hawkes process is a set of activation or excitation events, where each excitation has an exponential probability of exciting other events. Because of the exponential distribution it is useful to use a Laplace transformation to analyze the self exciting Hawkes processes.

\section{Data}
\label{data}
Telegram has been very successful in recent years \cite{dargahi2017analysis} as an instant messaging service (IM). While users can send text messages, photos, videos etc. to other users, the features that stand out in Telegram are its Channels, Groups, and the anonymity given to the users. We crawled specific channels in Telegram and logged conversational data using the publicly available API. Due to the nature of the conversations happening in these channels some users have been banned or deleted meaning that in some cases the crawler does not have access to the author of the message even though the message content exists in the thread. Moreover, when predicting the reply relations of a threaded structure the author of a message is quite important. We have taken the challenge of not using any meta data except for the time the post was created at.

\begin{figure}
  \centering
      \includegraphics[height=5cm]{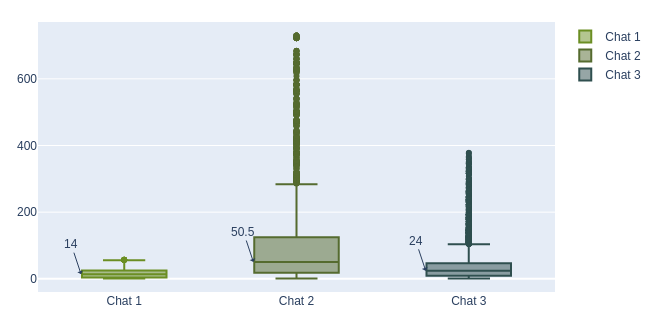}
      \caption{Distribution of length of the messages in each chat}
      \label{fig:boxplot}
\end{figure}

There is in interesting distribution of the message length in words. While most of the messages in each thread are short occasionally there are messages which has hundreds of words as well. These occasional messages are most of the time advertisements which seems to be selling or willing to buy things from users in the thread. Interestingly our encoded post embeddings are sensitive to the length of the message therefore it is easy to see these long messages as they seem to be outliers. For further reading we direct the readers to Section \ref{embeddings}.

We obtained three discussion threads and from here onward we shall call them `Chat 1', `Chat 2' and `Chat 3'. Each chat has messages, a unique identifier for any given message and a timestamp associated with it. Table \ref{datatable} is a descriptive table of how many messages and how long the discussion spans in each of these discussion threads. We have observed that there are short and long conversations present in the message threads, but conversations are not labeled.

\begin{table}[b]
  \centering
  \begin{tabular}{l  r  r}
  \hline
    \textbf{Thread} & \textbf{Number of messages} & \textbf{Time span (m)}\\
  \hline
  Chat 1 & 4832 & 1650.797\\
  Chat 2 & 3685 & 2415.881\\
  Chat 3 & 5597 & 2160.231\\
  \hline
  \end{tabular}
  \caption{Dataset description}
  \label{datatable}
\end{table}

\section{Model}
\label{model}
We create an LSTM based \cite{hochreiter1997lstm} model with a goal to learn vector representations to the posts in regards of their position of the thread and the language in the post itself. As discussed in the previous section to encode positional information we use the context window of surrounding posts. We created two paradigms for the context window, first has access to both messages which came before and after of the considering message. This allows the network to grasp positional information of the message from both sides, but this may not be the ideal in a real world setting. Claiming that a reply to a message will depend on the messages came before it, not the ones after. The second is when you take messages which came before a particular post and only them. We test both paradigms and evaluate results in the Section \ref{experiments}.

Following is a simple representation of the model and its objective function.

\[
l_i, W_i = \phi (p_i, C_i)
\] \[
loss\_value = similarity(l_i, W_i)
\]

Here ($p_i, C_i$) are post content and its context respectively. The model will process both these values and return a vector for the post and a vector for its context called ($l_i, W_i$) respectively. To learn the efficacy of ($l_i$) we employ a   similarity function, since both $l_i$ \& $W_i$ are vectors in the same dimension it is a matter of distance between the two vectors.

We learn time based info from the Hawkes/Laplace process based models from these vectors. This is required because a vanilla LSTM based models will not consider the time difference between two posts while it takes each post occurrence as events which happened in constant time. This is where the Hawkes process is important, it factors in where time difference is needed.

\begin{figure}[t!]
  \centering
  \begin{subfigure}[b]{0.3\linewidth}
  \includegraphics[width=\textwidth]{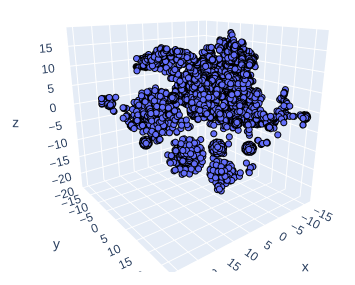}
    \caption{5$p_i$5} \label{atsne}
  \end{subfigure}
  \begin{subfigure}[b]{0.3\linewidth}
    \includegraphics[width=\textwidth]{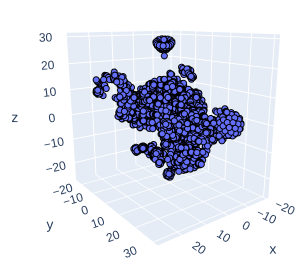}
    \caption{$4p_i$}\label{btsne}
  \end{subfigure}
  \begin{subfigure}[b]{0.3\linewidth}
    \includegraphics[width=\textwidth]{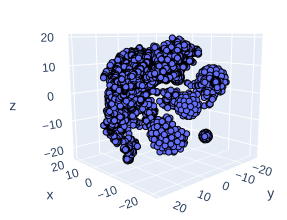}
    \caption{$10p_i$}\label{ctsne}
  \end{subfigure}
  \caption{Figure \ref{atsne} shows a first paradigm context window while Figures \ref{btsne} and \ref{ctsne} shows the second type, note that the original figure was rotated so that the clusters are evident in the illustration. In each illustration you can see dimensions $x, y$ \& $z$ which are calculated with the t-Stochastic Neighbor Embeddings \cite{van2008visualizing}}
\end{figure}

\section{Experiments}
\label{experiments}
Complete process of training the post embeddings to recreating the reply relations will be discussed here onward.
\subsection{Computing Post embeddings}
\label{embeddings}
First we shall discuss about the quality of the vector embeddings of the posts learned by our model. Linguistic and positional features were expected to learn as we trained these embeddings. To get a better understanding about our embeddings we compute t-Stochastic Neighbor Estimation and get 3-dimensional vectors for each of the embedding. We can use these vectors to visualize our embeddings and their quality. Using this it is trivial to examine the quality of the embeddings.

It is apparent in the figure \ref{ctsne} that obvious clusters exists in which when we explore more into them they are quite similar in language and the length of the post content. We used Agglomerative clustering and clustered into different number of clusters, and came into the above conclusion. It was also needed to change the size of the context window to see how that parameter will affect in the quality of the computed embeddings.

We transform naive Hawkes process with smoothed Laplace transformation.

\begin{figure}[t]
  \centering
  \begin{subfigure}[b]{0.49\linewidth}
    \includegraphics[width=\textwidth]{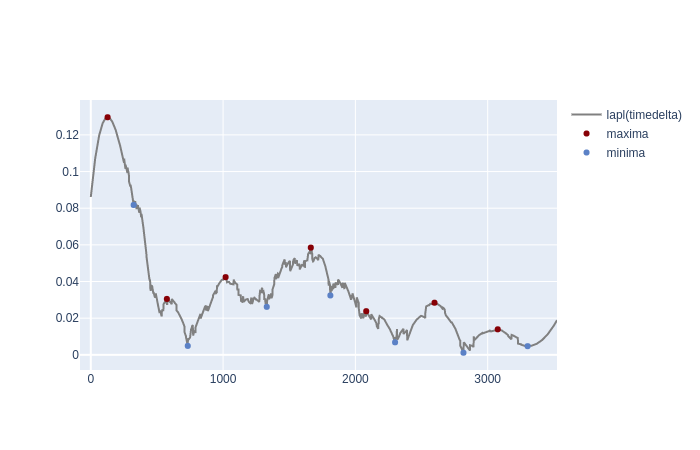}
    \caption{$Chat2$}\label{blap}
  \end{subfigure}
  \begin{subfigure}[b]{0.49\linewidth}
    \includegraphics[width=\textwidth]{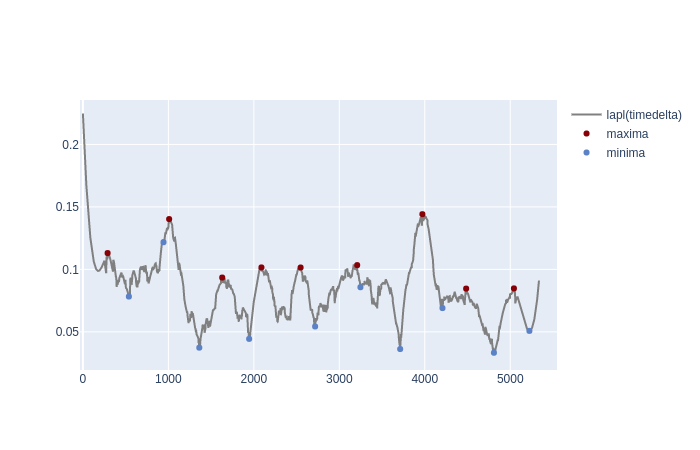}
    \caption{$Chat3$}\label{clap}
  \end{subfigure}
  \caption{This illustration shows the naive Hawkes process of Chats 2 and 3}
\end{figure}

\subsection{Algorithms}
\label{algorithm}
First drawing inspiration from \cite{wang2008recovering} we create a similarity matrix for the post embeddings in order to create a graph with edges as the   similarity. This will create a dense graph which is pruned using the Hawkes intensities. We calculate the average similarity score and then prune any edge with less than average score. This increases the number of conversations in the thread and allows the algorithm to find small conversations which happen inside bigger conversations.

\IncMargin{1em}
\begin{algorithm}[b]
\SetKwData{Ranges}{ranges}
\SetKwData{Embeddings}{embeddings}
\SetKwData{Pruned}{pruned}
\SetKwData{Sim}{similarityMatrix}
\SetKwData{Avg}{avg}
\SetKwFunction{Average}{averageScore}
\SetKwFunction{Zeros}{zeros}
\SetKwFunction{ }{ Similarity}
\SetKwFunction{In}{in}
\SetKwFunction{Add}{add}
\SetKwInOut{Input}{input}
\SetKwInOut{Output}{output}
\Input{\Ranges and \Embeddings}
\Output{Pruned Graph}

\Sim = \Sim(\Embeddings) \\
\Avg = \Average(\Sim) \\
\Sim[\Sim $<$ \Avg] = 0 \\
\Pruned = \Zeros(\Sim.shape) \\
\BlankLine
\For{(min, max) \In \Ranges}{
\Pruned[min:max, min:max] = \Sim[min:max, min:max]
}
\caption{Pruning Algorithm}
\label{algo_1pruning}
\end{algorithm}
\DecMargin{1em}

This algorithm will yield the pruned graph structure. It is satisfactory to assume that a post with an earlier time stamp will never be a child to a post which came after that. The resulting graph is a directed and sparse structure which represents the thread. Post processing of the graph will give us multiple discussion threads with one root message for each thread. This first run of the pruning algorithm will help to determine the root messages of the threads, however this tree like graph is has isolated sub graphs and each of these are shallow having maximum depth one, we can use another algorithm to compute the path of the message structure to reconstruct the threads parent post structure essentially creating a sparse pruned representation of the same graph.

\begin{figure}
  \centering
  \includegraphics[height=5cm]{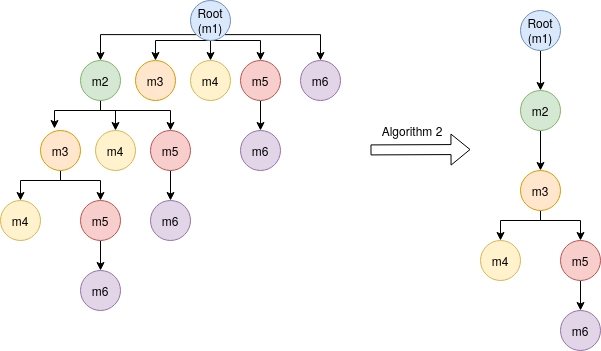}
  \caption{Algorithm 2 input and output}
\end{figure}

Before the second stage of the pruning algorithm, the graph indicates edges from all the children from a parent. This means that a parent can have edges to its grand children. Goal of Algorithm 2 is to increase the depth of the isolated sub graphs, essentially this will prune the repetitive nodes making it a sensible discussion structure.

\IncMargin{1em}
\begin{algorithm}[t]
\SetKwData{Graph}{graph}
\SetKwData{Child}{child}
\SetKwData{Children}{children_}
\SetKwData{Pruned}{pruned}
\SetKwData{Visited}{visited}
\SetKwFunction{Remove}{remove}
\SetKwFunction{Addchildren}{setchildren}
\SetKwFunction{Children}{children}
\SetKwFunction{In}{in}
\SetKwFunction{Add}{add}
\SetKwInOut{Input}{input}
\SetKwInOut{Output}{output}
\Input{Graph pruned (\Pruned) with Algorithm 1}
\Output{Graph after removing extra linkages}

\For{$node$ \In \Graph}{
  \Visited.\Add($node$)\\
  \Children = \Graph.\Children(node)\\
  \Pruned.\Addchildren(\Children) \\

  \For{$itm$ \In \Visited}{
    \If{$node$ == $itm$}{
      continue\\
    }
    \For{\Child \In \Pruned.\Children(node)}{
      \If{\Child \In \Pruned.\Children(itm)}{
        \Pruned.\Children(itm).\Remove(\Child)
      }
    }
  }
}
\caption{Further prunning - thinning the graph}
\label{algo_2pruning}
\end{algorithm}
\DecMargin{1em}

\section{Conclusions}
\label{conclusions}
This article presents a disentanglement method for threaded discussions where the reply relation will be discovered using a position encoded and linguistic embedding with the help of a graph pruning algorithm. We have shown that our method built on top of most of the literature shared herewith have been able to increment the performance of them.

While our method removes two phases of training similar to \cite{wang2008recovering} it is also not a complete end to end process. There is a separate entity (neural network) to create the embeddings, the graph structure is another entity with its own properties and parameters which needs adjusting. It is true that we have fine tuned our parameters for the data but for further scaling and managing it would be much easier to have one system which needs to be trained or parameter tuned once. We are using a top down construction of a graph and when one node is visited, the algorithm does not evaluate to the nodes above it to see whether it is attached to others, meaning that if a message is replying to multiple parent messages, we do not have the means to discover it yet.

\bibliography{refs}
\bibliographystyle{plainnat}
\end{document}